\pdfoutput=1

\documentclass[11pt]{article}

\usepackage[]{acl}

\usepackage{times}
\usepackage{latexsym}
\usepackage{graphicx}

\usepackage[T1]{fontenc}

\usepackage[utf8]{inputenc}

\usepackage{microtype}
\usepackage{graphicx}
\usepackage{bm}
\usepackage{amsmath}
\title{Logic Traps in Evaluating Attribution Scores}

\author{
    Yiming Ju\textsuperscript{\rm 1,2},
    Yuanzhe Zhang\textsuperscript{\rm 1,2},
    Zhao Yang\textsuperscript{\rm 1,2},   \\
    \textbf{Zhongtao Jiang\textsuperscript{\rm 1,2},}
    \textbf{Kang Liu\textsuperscript{\rm 1,2,3},} 
    \textbf{Jun Zhao\textsuperscript{\rm 1,2},}
    \\
    \textsuperscript{\rm 1} National Laboratory of Pattern Recognition, Institute of Automation, CAS, Beijing, China \\
    \textsuperscript{\rm 2} School of Artificial Intelligence, University of Chinese Academy of Sciences, Beijing, China \\
    \textsuperscript{\rm 3} Beijing Academy of Artificial Intelligence, Beijing, 100084, China \\
    \texttt{\{yiming.ju, yzzhang, zhao.yang\}@nlpr.ia.ac.cn} \\
    \texttt{\{zhongtao.jiang, kliu, jzhao\}@nlpr.ia.ac.cn} \\
}

\begin{document}
\maketitle
\begin{abstract}
Modern deep learning models are notoriously opaque, which has motivated the development of methods for interpreting how deep models predict.
This goal is usually approached with attribution method, which assesses the influence of features on model predictions. 
As an explanation method, the evaluation criteria of attribution methods is how accurately it reflects the actual reasoning process of the model (faithfulness). Meanwhile, since the reasoning process of deep models is inaccessible, researchers design various evaluation methods to demonstrate their arguments.
However, some crucial logic traps in these evaluation methods are ignored in most works, causing inaccurate evaluation and unfair comparison.
This paper systematically reviews existing methods for evaluating attribution scores and summarizes the logic traps in these methods.
We further conduct experiments to demonstrate the existence of each logic trap.
Through both theoretical and experimental analysis, we hope to increase attention on the inaccurate evaluation of attribution scores. 
Moreover, with this paper, we suggest stopping focusing on improving performance under unreliable evaluation systems and starting efforts on reducing the impact of proposed logic traps.

\end{abstract}

\section{Introduction}

\begin{figure}[htb]
	\centering
	\includegraphics[width=0.99\linewidth]{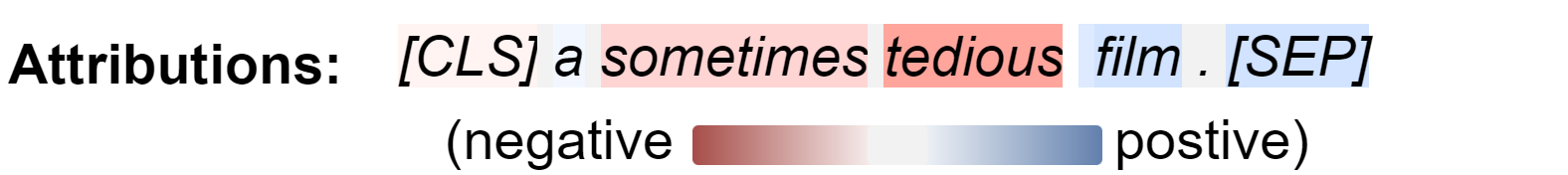}
	\caption{An example of attribution explanations, which assesses the influence of each token on the predictions of a binary sentiment classification task. The saturation of the colors signifies the magnitude of the influence.
	}
	\label{attribution}	
\end{figure}

The opaqueness of deep models has grown in tandem with their power   \citep{doshi2017towards}, which has motivated efforts to interpret how these black-box models work \citep{sundararajan2017axiomatic, belinkov2019analysis}. 
Post-hoc explanation aims to explain a trained model and reveal how the model arrives at a decision \citep{jacovi2020towards, molnar2020interpretable}. 
This goal is usually approached with attribution method, which assesses the influence of features on model predictions as shown in Figure \ref{attribution}.
Recent years have witnessed an increasing number of attribution methods being developed. For example, Erasure-based method calculate attribution scores by measuring the change of output after removing target features \citep{li2016understanding, feng2018pathologies, chen2020generating};
Gradient-based method uses gradients to study the influence of features on model predictions \citep{sundararajan2017axiomatic, wallace2019allennlp, hao2020self};
Meanwhile, these methods also received much scrutiny, arguing that the generated attribution scores are fragile or unreliable
\citep{alvarez2018robustness, pruthi2019learning, wang2020gradient, slack2020fooling}. 

\emph{As an explanation method, the evaluation criteria of attribution methods should be how accurately it reflects the true reasoning process of the model (faithfulness), not how convincing it is to humans (plausibility)}
\citet{jacovi2020towards}.
Meanwhile, since the reasoning process of deep models is inaccessible, researchers design various evaluation methods to support their arguments, some of which appear valid and are widely used in the research field.
For example, \emph{meaningful perturbation} is used for making comparison in many works \citep{samek2016evaluating, chen2018shapley, deyoung2019eraser, chen2020generating, kim2020interpretation}.
The philosophy of \emph{meaningful perturbation} is simple, i.e., modifications to the input instances, which are in accordance with the generated attribution scores, can bring about significant differences to model predictions if the attribution scores are faithful to the target system.

However, some crucial logic traps existing in these evaluation methods are ignored in most works, causing inaccurate evaluation and unfair comparison.
For example, we found that we can manipulate the evaluation results when using \emph{meaningful perturbation} to make comparisons:
by choosing the modification strategy, we can assign any of the three candidate attribution methods as the best method.
The neglect of these traps has damaged the community in many aspects:
First, the existence of logic traps will lead to an inaccurate evaluation and unfair comparison, making the conclusions unreliable;
Second, the wide use of evaluation metrics with logic traps brings pressure to newly proposed works, requiring them to compare with other works using the same metrics;
Last, the over-belief in existing evaluation metrics encourages efforts to propose more accurate attribution methods, notwithstanding the evaluation system is unreliable.

In this paper, we systematically review existing methods for evaluating attribution scores and categorize them into classes.
We summarize the logic traps in these methods and further conduct experiments to demonstrate the existence of each logical trap.
Though strictly accurate evaluation metrics for attribution methods might be a “unicorn” which will likely never be found, we should not just ignore logic traps in existing evaluation methods and draw conclusions recklessly.
Through both theoretical and experimental analysis, we hope to increase attention on the inaccurate evaluation of attribution scores. 
Moreover, with this paper, we suggest stopping focusing on improving performance under unreliable evaluation systems and starting efforts on reducing the impact of proposed logic traps.

\section{Evaluation Methods and Corresponding Logic Traps}

\subsection{Part \uppercase\expandafter{\romannumeral1}}

\noindent
\textbf{Evaluation 1: Using Human Annotated Explanations As the Ground Truth}

Evaluation 1 verifies the validity of the attribution scores by comparing them with the human problem-solving process. 
In this evaluation, works (e.g., \citet{murdoch2018beyond, kim2020interpretation, sundararajan2017axiomatic}) often give examples consistent with human understandings to demonstrate the superiority of their proposed method.
For example, as shown in Table \ref{Heat Map}, \citet{murdoch2018beyond}  shows heat maps for a yelp review generated by different attribution techniques.
They argue that the proposed method: Contextual decomposition, is better than others because only it can identify \emph{`favorite'} as positive and \emph{`used to be'} as negative, which is consistent with human understandings. 

\begin{table}[t!]
\resizebox{0.48\textwidth}{!}{
\centering
\begin{tabular}{|l|c|}
\hline
Method &  Heat Map \\
\hline
Leave One Out & \textcolor[RGB]{200,0,0}{\emph{used}}  \textcolor[RGB]{181,144,41}{\emph{to}} \textcolor[RGB]{181,144,41}{\emph{be}} \textcolor[RGB]{181,144,41}{\emph{my}} \textcolor{orange}{\emph{favorite}} \\
\hline
Integrated gradients & \textcolor[RGB]{0,0,200}{\emph{used}}  
\textcolor[RGB]{0,90,0}{\emph{to}} 
\textcolor[RGB]{181,144,41}{\emph{be}}
\textcolor[RGB]{181,144,41}{\emph{my}} \textcolor[RGB]{181,144,41}{\emph{favorite}} \\
\hline
Contextual decomposition & \textcolor[RGB]{200,0,0}{\emph{used}}  
\textcolor[RGB]{200,0,0}{\emph{to}} 
\textcolor[RGB]{200,0,0}{\emph{be}}
\textcolor[RGB]{0,90,0}{\emph{my}} \textcolor[RGB]{0,0,200}{\emph{favorite}} \\
\hline
\multicolumn{2}{c}{\small Legend: \textcolor[RGB]{200,0,0}{Very Negative} \textcolor{orange}{Negative} \textcolor[RGB]{181,144,41}{Neutral} \textcolor[RGB]{0,90,0}{Positive} \textcolor[RGB]{0,0,200}{Very Positive}} \\
\end{tabular}
}
\caption{Heat maps for a portion of a yelp review generated by different attribution techniques.
The example and results are taken from \citet{murdoch2018beyond}.
}
\label{Heat Map}
\end{table}

Furthermore, resorting to human-annotated explanations, works can also evaluate attribution methods quantitatively in evaluation 1.
For example, the SST-2 \citep{socher2013recursive} corpus provides not only sentence-level
labels, but also five-class word-level sentiment tags
ranging from very negative to very positive. 
Thus, many works \citep{lei2016rationalizing, li2016understanding, tsang2020does, kim2020interpretation} perform quantitative evaluation of attribution scores by comparing them with the word-level tags in SST-2.

~\\
\noindent
\textbf{Logic Trap 1: The decision-making process of neural networks is not equal to the decision-making process of humans.}

First, we cannot completely deny the rationality of evaluation 1.
Since many attribution methods work without any human-annotated information, such as erasure-based and gradient-based methods, the similarity between human-annotated explanations and generated attribution scores can be seen as drawing from the reasoning process of target models.
However, since the deep model often rely on unreasonable correlations, even when producing correct predictions, 
attribution scores preposterous to humans may reflect the reasoning process of the deep model faithfully. 
Thus we cannot deny the validity of an attribution score through its inconsistency to human-annotated explanations and cannot use human-annotated explanations to conduct a quantitative evaluation.

~\\
\noindent
\textbf{Experiment 1:}

In experiment 1, we give an example to demonstrate that the model might rely on correlations inconsistent with human understandings to get the prediction: 
though trained with questions, a question-answering model could maintain the same prediction for a large ratio of samples when the question information is missing, which is obviously different from humans.

We experiment on RACE \citep{lai2017race}, a large-scale question-answering dataset. As shown in Table \ref{RACE}, RACE requires the model to choose the right answer from candidate options according to the given question and document. 

\begin{table}[htb]
\resizebox{0.48\textwidth}{!}{
\centering
\begin{tabular}{|p{7.8cm}|}
\hline
\small \textcolor{blue}{\emph{\textbf{Document:}}} \emph{``...Many people optimistically thought industry awards for better equipment
would stimulate the production of quieter
appliances. It was even suggested that noise from
building sites could be alleviated ...''} \\

\small \textcolor{blue}{\emph{\textbf{Question:}}} \emph{What was the author’s attitude towards the industry awards for quieter?}

\small \textcolor{blue}{\emph{\textbf{Options:}}} 
\ \ \ A. suspicious \ \ \ \ \ \ \ B. positive \\
\small \ \ \ \ \ \ \ \ \ \ \ \ \ \ \ \ \ \ C. enthusiastic 
\ \ \ \ \  D. indifferent \\
\hline
\end{tabular}
}
\caption{An sample taken from RACE dataset.
}
\label{RACE}
\end{table}

\begin{figure}[htb]
	\centering
	\includegraphics[width=0.85\linewidth]{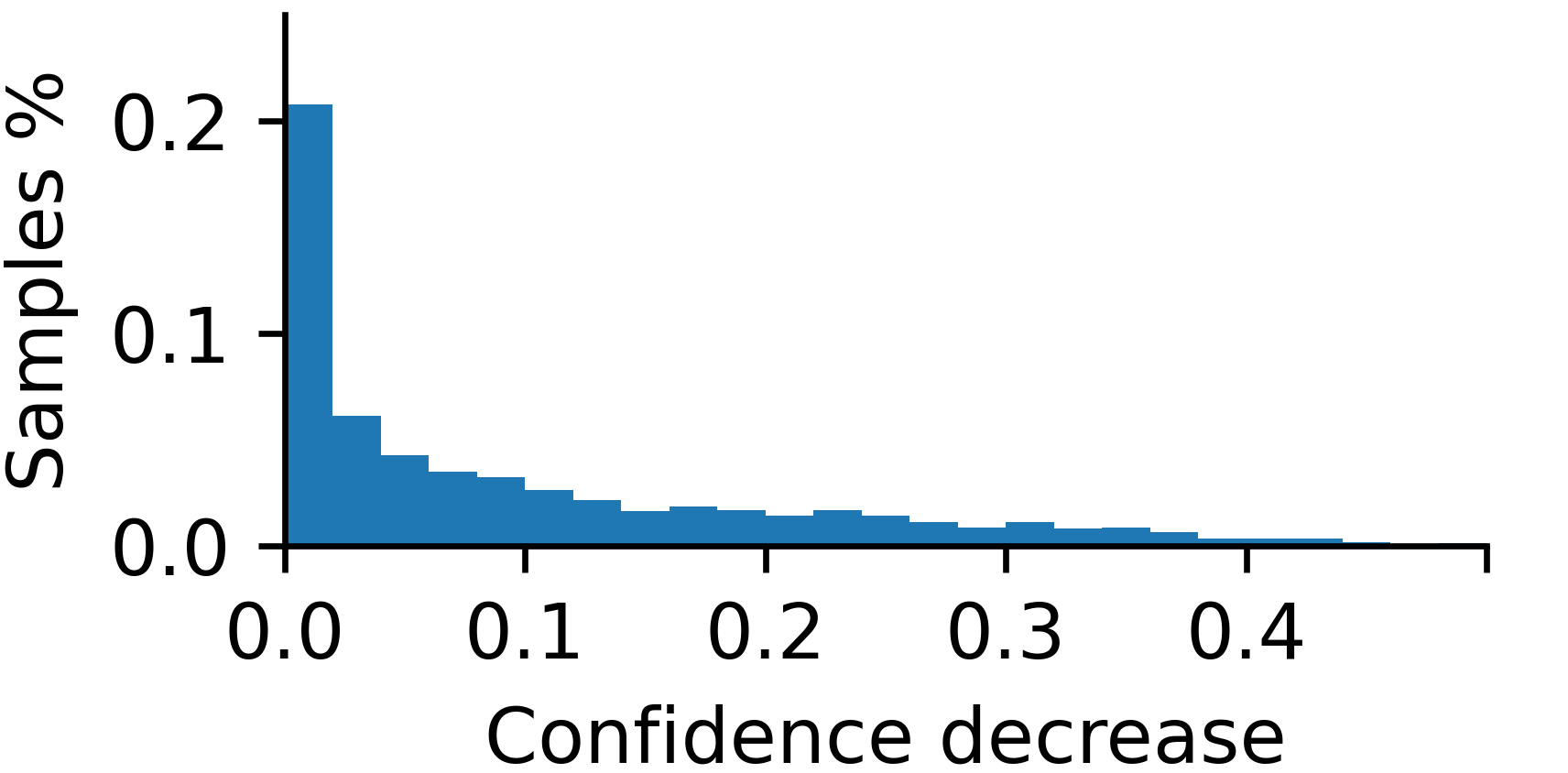}
	\caption{Confidence decrease in unchanged samples.
	}
	\label{RACEresult}	
\end{figure}

We first train a model with BERT$_{base}$ \citep{devlin2018bert} as encoder\footnote{Our implementations of experiment 1 and experiment 2 are based on the Huggingface’s transformer model hub (\url{https://github.com/huggingface/transformers}), and we use its default model architectures without change for corresponding tasks. \label{web}} with questions,
and achieve 65.7\% accuracy on the development set.
Then, we replace the development set questions with empty strings and feed them into the trained model.
Surprisingly, the trained MRC model maintained the original prediction on 64.0\% of the test set samples (68.4\% on correctly answered samples and 55.4\% on wrongly answered samples).
Moreover, we analyze the model confidence change in these unchanged samples, where the probability on the predicted label is used as the confidence score.
As shown in Figure \ref{RACEresult}, most of the samples have confidence decrease smaller than 0.1, demonstrating question information are not essential for the model to get predictions in these samples.

Since question information is usually crucial for humans to answer the question, attribution scores faithfully reflect the reasoning process of this model may be inconsistent with human annotations.
Thus, it is improper to use human-annotation explanations as the ground truth to evaluate attribution methods.

\subsection{Part \uppercase\expandafter{\romannumeral2}}

~\\
\noindent
\textbf{Evaluation 2: Evaluation Based on Meaningful Perturbation}

Most existing methods for quantitatively evaluating attribution scores can be summarized as evaluations based on  \emph{meaningful perturbation}.
The philosophy of \emph{meaningful perturbation} is simple, i.e., modifications to the input instances, which are in accordance with the generated attribution scores, can bring about significant differences to the target model's predictions if the attribution scores are faithful to the target model.

\begin{figure}[htb]
	\centering
	\includegraphics[width=0.55\linewidth]{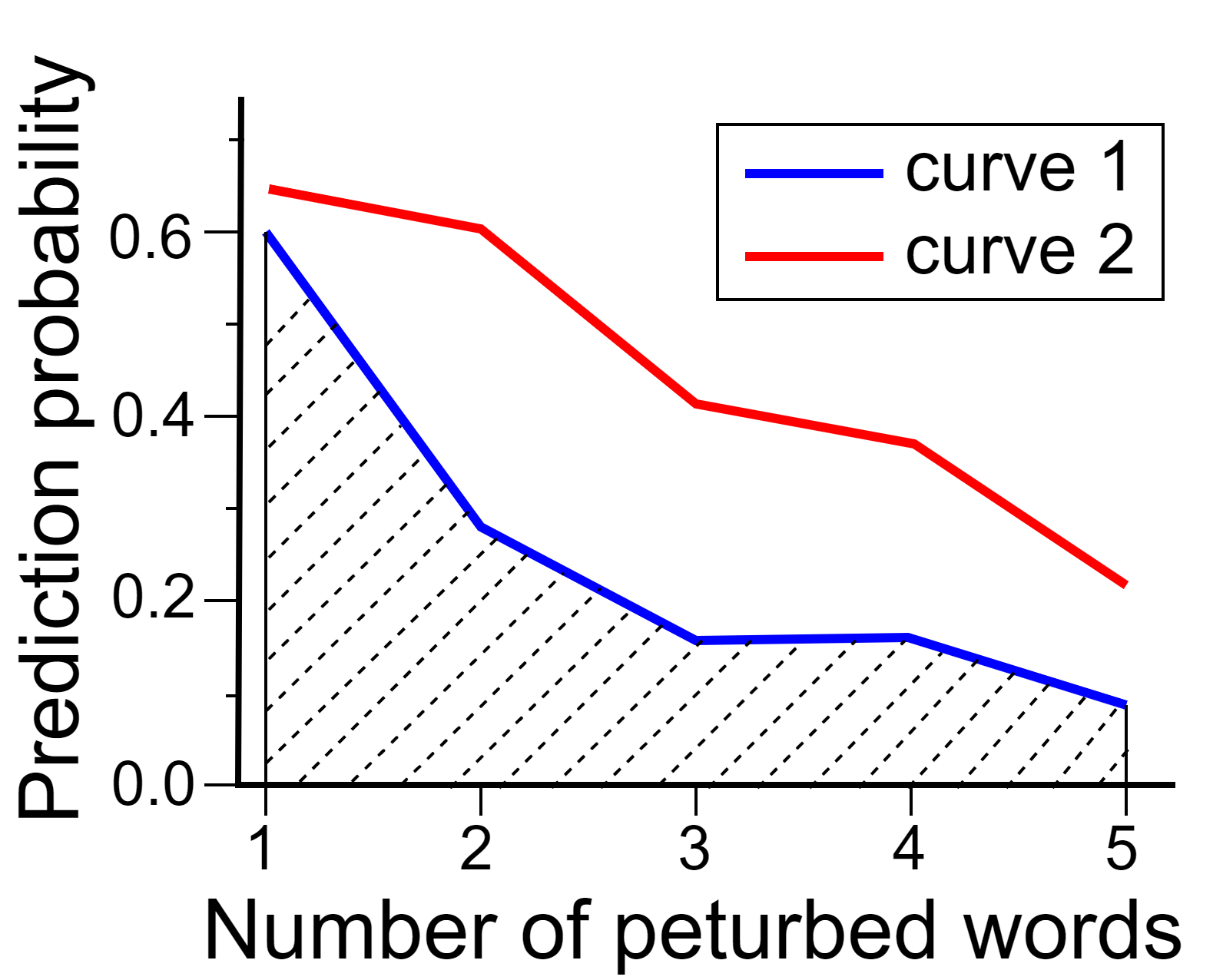}
	\caption{AUC evaluation metric. The smaller area under the curve, the better the result.}
	\label{AUC}	
\end{figure}

For example, \citet{samek2016evaluating, nguyen2018comparing, chen2020learning} use the area over the perturbation curve (AOPC) \citep{samek2016evaluating} as evaluation metrics.
Specifically, given the attribution scores of a set of features, AOPC(k) modifies the top k\% features and calculates the average change in the prediction probability as follows,
\[
AOPC(K) = \frac{1}{N}\sum_{i=1}^{N}\left \{p(\hat{y}|x_{i}) - p(\hat{y}|\tilde{x}_{i}^{(k)}) \right \}
\]
where $\hat{y}$ is the predicted label, $N$ is the number of examples, $p(\hat{y}|·)$ is the probability on the predicted class, and $\tilde{x}_{i}^{(k)}$ is modified sample. Higher AOPCs is better, which means that the features chosen by attribution scores are
more important;
\citet{feng2018pathologies, petsiuk2018rise, kim2020interpretation} use area under the curve (AUC) to evaluate attribution scores. As shown in Figure \ref{AUC}, AUC plots a prediction probability curve about modified feature numbers where features are modified in order of attribution scores. The argument is if attribution scores are faithful, then the curve will drop rapidly, resulting in a small area under a curve. 

Besides these works, a lot of works \citep{shrikumar2017learning, chen2018shapley, nguyen2018comparing, deyoung2019eraser, chen2020generating, hao2020self, jiang-etal-2021-alignment} use similar metrics to perform evaluation and comparisons.
The main difference between evaluation metrics in these works is the difference in the modification strategy.
For example, to evaluate word-level attribution scores for SST-2, \citet{chen2020generating} uses deleting tokens as modification while
\citet{kim2020interpretation} uses replacing tokens with tokens sampled from the distribution inferred by BERT.

~\\
\noindent
\textbf{Logic Trap 2: Using an attribution method as the ground truth to evaluate the target attribution method.}

Evaluation methods based on \emph{meaningful perturbation} can be seen as an attribution method too.
For example, AOPC(k), which assesses the importance of k\% features, can be seen as an attribution method calculating an attribution score for k\% features. 
Specifically, when using deleting tokens as modification and narrowing the k\% to one token,
AOPC(k) degenerates into the basic attribution method: \textbf{leave-one-out} \citep{li2016understanding}. 
Thus, evaluation 2 uses an attribution method as the ground truth to evaluate the target attribution method,
which measures the similarity between two attribution methods instead of faithfulness.

Since \emph{meaningful perturbation} assesses the importance of features by calculating output change after modifications, its results 
are mainly depend on how to conduct the modifications, which means 
different modification strategies might lead to different evaluation results.
Evaluation 2 is widely used to compare attribution methods in the research field. Accordingly, the neglect of logic trap 2 has led to a high risk of unfair comparisons and unreliable conclusions.

~\\
\noindent
\textbf{Experiment 2:}

In experiment 2, we give an example of unfair comparisons in evaluation 2: the more similar the target attribution method to the modification strategy, the better the evaluation results. 
Specifically, by modifying the modification strategies in APOC and AUC, we can assign any of the three candidate attribution methods as the best method.
We conduct experiments on on widely used SST-2 task of the GLUE benchmark \citep{wang2018GLUE}),
and use BERT$_{base}$ as encoder to build the target model\textsuperscript{\ref {web}} (achieve 86.4\% accuracy).

\begin{figure}[t]
	\centering
	\includegraphics[width=0.95\linewidth]{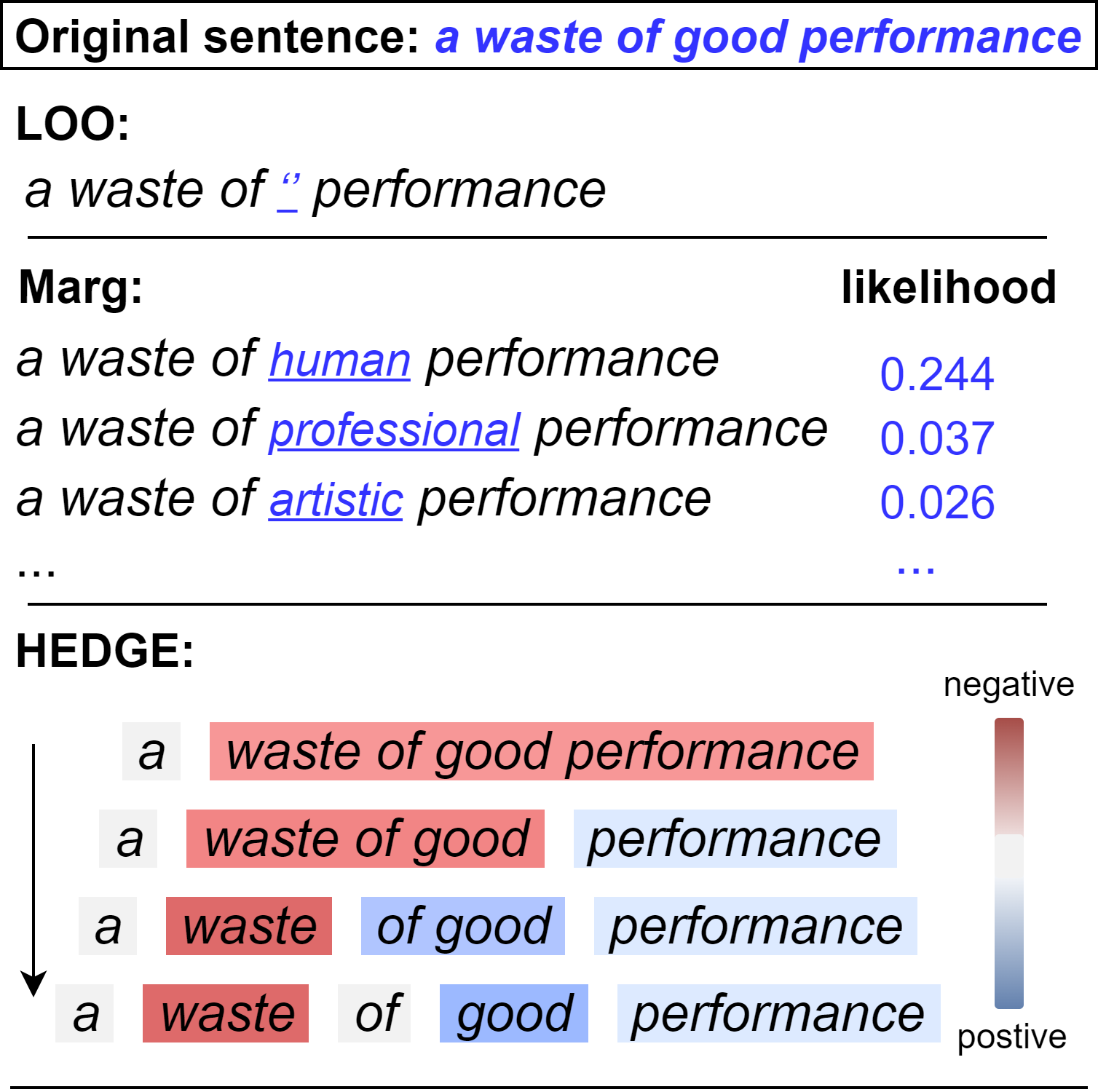}
	\caption{The overview of LOO, Marg and HEDGE.}
	\label{method12}	
\end{figure}

~\\
\noindent
\textbf{Attribution Methods \ \ }
We experiment with three attribution methods:
leave-one-out (LOO) \citep{li2016understanding}, HEDGE \citep{chen2020generating} and Marg \citep{kim2020interpretation}.
The schemes of these attribution methods are shown in Figure \ref{method12}, LOO assign attribution scores to the target word \emph{`good'} by deleting it from the sentence and observing change in the model predictions;
Marg marginalizes the target word \emph{`good'} out considering the likelihoods of all candidate words, which uses BERT to measure the likelihoods of candidate words to replace the target word;
HEDGE builds hierarchical explanations by recursively
detecting the weakest interactions and then dividing large text spans into smaller ones.
HEDGE assign attribution scores to spans by using '[PAD]' token to replace other words in a sentence and measuring how far the prediction is to the prediction boundary.

\begin{table*}[htb]
\centering
\begin{tabular}{|l|c|c|c|c|c|c|}
\hline
Method/Metric & AOPC$_{del}$ $\bm{\uparrow}$   & AUC$_{rep}$ $\bm{\downarrow}$
& AOPC$_{rep}$ $\bm{\uparrow}$ & AUC$_{del}$ $\bm{\downarrow}$
& AOPC$_{pad}$  $\bm{\uparrow}$ & AUC$_{pad}$  $\bm{\uparrow}$ \\
\hline
LOO 
& \textbf{0.541} &   0.666  
& 0.378 &   \textbf{0.526}     
& 0.935 &   0.896  \\
HEDGE 
& 0.466 & 0.702
& 0.324  &   0.638  
& \textbf{0.978}  &  \textbf{0.984}\\
Marg 
 & 0.477  &  \textbf{0.617}  
& \textbf{0.391}  &  0.588     
 & 0.928  &  0.874\\
\hline
\end{tabular}
\caption{Evaluation results of three attribution methods. $\bm{\uparrow} /\bm{\downarrow}$ refers to higher / lower scores are better. ${del}$, ${rep}$, and ${pad}$ refer to different modification strategies in the evaluation metrics.}
\label{m122}
\end{table*}

~\\
\noindent
\textbf{Evaluation metrics and Results \ \ }
We first evaluate three attribution methods with metrics drawn from Marg and HEDGE papers.
Marg uses AUC as evaluation metrics and modifies words by gradually replacing them with a token sampled from the distribution inferred by BERT, denoted as AUC$_{rep}$;
HEDGE uses AOPC as evaluation metrics and modifies words by deleting them directly, denoted as AOPC$_{del}$.
Both papers modify 20\% of words in the sentence.
The results are shown in Table \ref{m122}.
As shown in Table \ref{m122}, Marg performs best in AUC$_{rep}$ while
LOO performs best in AOPC$_{del}$.
Since the modification strategy of AOPC$_{del}$ is consistent with LOO, and that of AUC$_{rep}$ is most similar to Marg, 
the evaluation results are consistent with the inference in logic trap 2: \emph{the more similar the target evaluated method to the modification strategy, the better the evaluation results.}

~\\
\noindent
\textbf{Manipulate Evaluation Results \ \ }
We further conduct ablation experiments by changing the modification strategies in AOPC$_{del}$ and AUC$_{rep}$.
Concretely, we switched perturbing strategy in AOPC$_{del}$ and AUC$_{rep}$ and get new evaluation metrics: AOPC$_{rep}$ and AUC$_{del}$. 
As shown in Table \ref{m122}, different from the initial results, Marg performs best in APOC metric while LOO performs best in AUC metric.
The opposite results demonstrate that evaluation results mainly depend on the modification strategies, and we can manipulate evaluation results by changing them. 
Moreover, we note that HEDGE performs worst in all four evaluation metrics. Thus, we further customize the modification strategy to HEDGE's advantage: padding unimportant features according to the attribution scores, denoted as AOPC$_{pad}$ and AUC$_{pad}$.
Not surprisingly, results in Table \ref{m122} show that HEDGE perform best in customized metrics.

~\\
\noindent
\textbf{Summarization \ \ }
Because of the existence of logic trap 2, we can manipulate the evaluation results in evaluation 2 by changing the modification strategies, assigning any of the three candidate attribution methods as the best method.
In fact, because we cannot simply assign a modification strategy as faithful, we should not use evaluation 2 to quantitatively evaluate attribution scores and make comparisons.
Since the wide use of evaluation 2, the neglect of logic trap 2 has negatively impacted the research field for a long time.
First, it brings a risk of unfair comparisons: works can customize evaluation metrics to their advantage and thus achieve the best performance.
Second, the wide use of evaluation 2 also brings pressure to new proposed works, forcing them to make comparisons to others in such evaluation.

\subsection{Part \uppercase\expandafter{\romannumeral3}}

~\\
\noindent
\textbf{Evaluation 3: Disprove Attribution Methods by Examining the Consistency of Attribution Scores}

In this evaluation, works evaluate attribution methods by examining the consistency of attribution scores for similar inputs. 
The philosophy of Evaluation 3 is that semantically similar inputs which share the same model predictions should have similar attribution scores if the attribution method is reliable.
Evaluation 3 is often used to disprove the effectiveness of attribution methods by searching for counterexamples.

\begin{figure}[htb]
	\centering
	\includegraphics[width=0.99\linewidth]{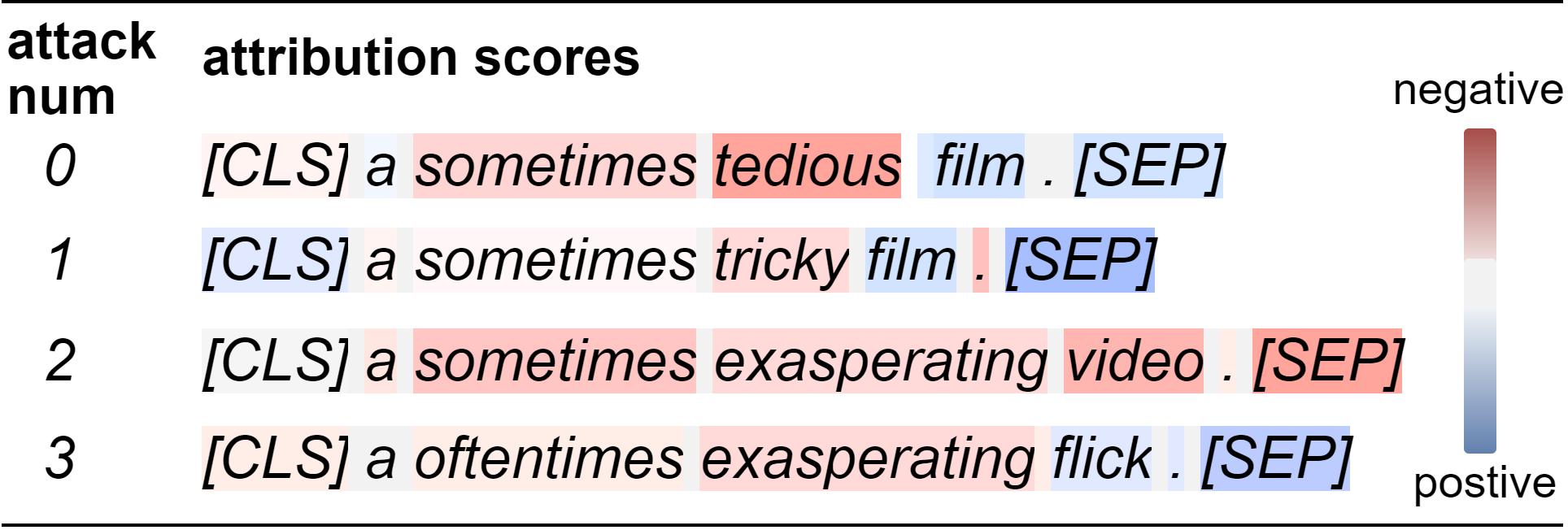}
	\caption{Examples taken from ExplainFooler \citep{sinha2021perturbing}, which attacks attribution methods by searching adversarial sentences with different attribution scores. \emph{attack num} refers to the number of replaced words.} 
	\label{attack}	
\end{figure}

For example, ExplainFooler \citep{sinha2021perturbing} attacks Integrated Gradients and \citep{sundararajan2017axiomatic} and LIME \citep{sundararajan2017axiomatic}, which are two popular attribution methods in NLP, by searching adversarial sentences with different attribution scores.
As shown in Figure \ref{attack}, these adversarial sentences are semantically similar to the original sentence and share the same model predictions.
However, the attribution scores of these sentences are very different from that of the original sentence. \citet{sinha2021perturbing} observes the rank order correlation drops by over 20\% when less than 10\% of words are changed on average and thus draws the conclusion that Integrated Gradients and LIME are fragile. 

A lot of works \citep{alvarez2018robustness, kindermans2019reliability, ghorbani2019interpretation, ding2021evaluating, sinha2021perturbing} use evaluation 3 to examine the validity of existing attribution methods. For example, \citet{ghorbani2019interpretation} argues that interpretations of neural networks are fragile by showing that systematic perturbations can lead to different interpretations without changing the label; \citet{alvarez2018robustness} argues that a crucial property that attribution methods should satisfy is robustness to local perturbations of the input.

~\\
\noindent
\textbf{Logic Trap 3: The change in attribution scores maybe because the model reasoning process is really changed rather than the attribution method is unreliable.}

When solving similar samples like those shown in Figure \ref{attack}, humans tend to use similar reasoning methods.
However, deep models are not as robust enough as humans and often rely on unreasonable correlations. 
Semantically similar texts often cause different reasoning processes in deep models.
For example, it is well known that deep models are vulnerable to adversarial samples \citep{goodfellow2014explaining, papernot2016limitations}.
By deliberately adding some subtle interference that people cannot detect to the input sample, the target model will give a different prediction with high confidence.
The success in adversarial attacks on deep models demonstrates similar inputs for humans can share very different reasoning processes in deep models.

The main difference between attribution-attacking methods and model-attacking is that attribution-attacking methods require the model to give the same prediction for adversarial samples. 
However, giving the same prediction is very weak to constraint model reasoning
because deep models have compressed the complicated calculation process into limited classes in the prediction.
For example, there is always half probability of giving the same prediction for a binary classification task even with totally random reasoning.
Thus, it is no surprise that attribution-attacking methods can find adversarial samples which share the same prediction label to the original sample yet have different attribution scores.

\begin{figure}[htb]
	\centering
	\includegraphics[width=0.9\linewidth]{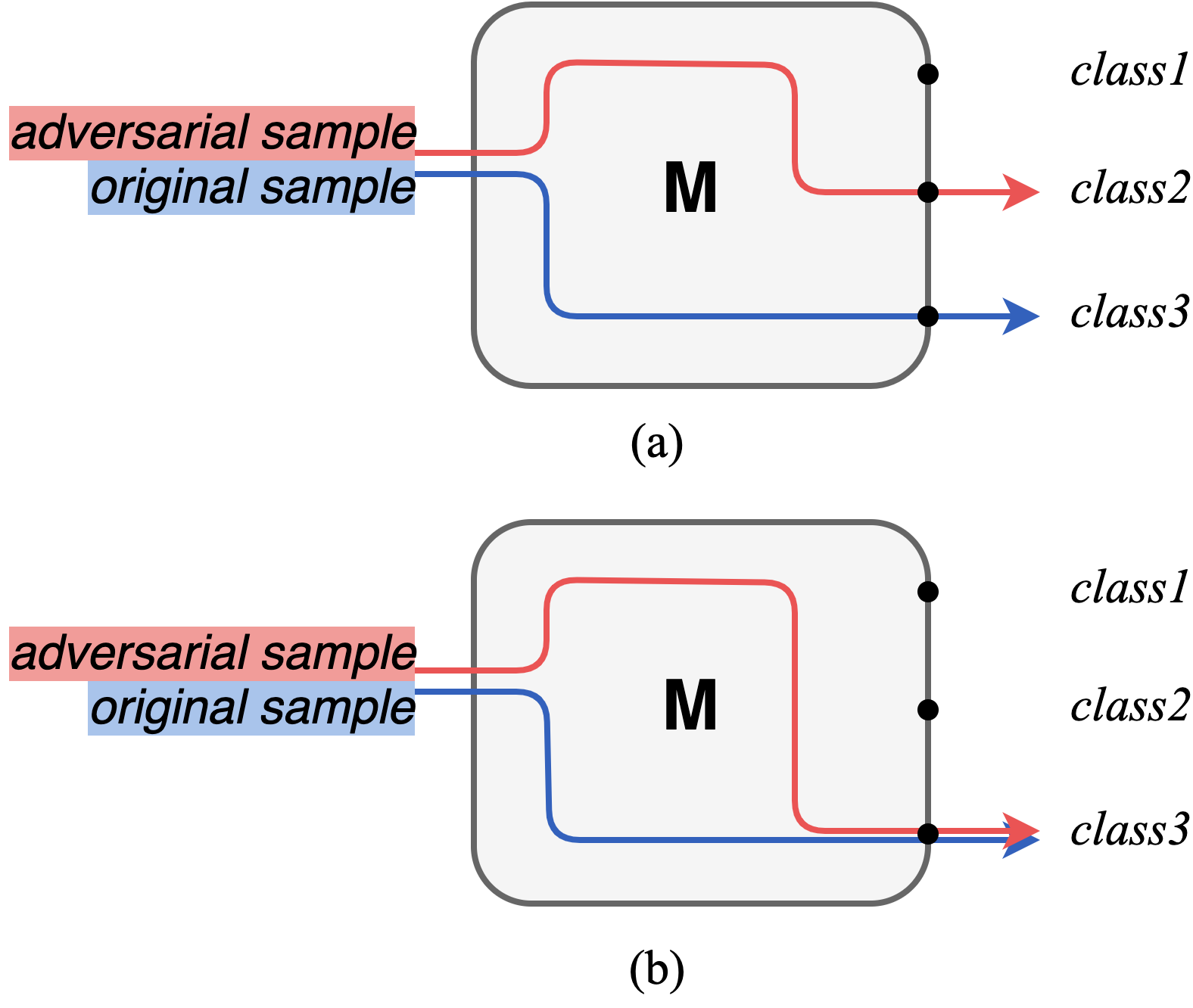}
	\caption{
    We use lines connecting inputs and outputs to represent the model reasoning process.
	(a) is a successful attack on the target model while (b) might be regarded as a successful attack on attribution methods, falling into the logic trap 3.
	}
	\label{F5}	
\end{figure}

The logic trap in evaluation 3 is that the change in attribution scores may be because the model reasoning process is really changed rather than the attribution method is unreliable.
As shown in Figure \ref{F5}. (b), an attribution method should generate different attribution scores for the original and adversarial samples if it faithfully reflects the model reasoning. However, it will be regarded as 
fragile or unreliable in evaluation 3.
Unfortunately, existing works ignore this logic trap and propose various methods to attack attribution methods.
Since the high susceptibility of deep models to adversarial samples, not surprisingly, all of these works get the same conclusion: existing attribution methods are fragile or unreliable.

~\\
\noindent
\textbf{Experiment 3:}

In experiment 3, we demonstrate that deep models can assign the same label to semantically similar samples yet use different reasoning.
We experiment on widely used SST-2 and MNLI tasks of the GLUE benchmark \citep{wang2018GLUE}).
MNLI requires the model to predict whether the premise entails the hypothesis, contradicts it, or is neutral.

~\\
\noindent
\textbf{Model \ \ }
Since the attribution methods are defaulted as unreliable in evaluation 3, we cannot use existing attribution methods to judge whether the model reasoning is different.
To solve the problem, we use a two-stage model framework, where the model first extracts a subset of inputs and gives prediction based only on the subset information.
This way, we can observe whether the model reasoning is changed from the chosen subset, i.e., different subsets means the model chooses to use different information to make the final decision\footnote{Note that similar subsets are regarded as a necessary condition rather than a sufficient condition for the similar model reasoning process.}.

\begin{figure}[t!]
	\centering
	\includegraphics[width=0.8\linewidth]{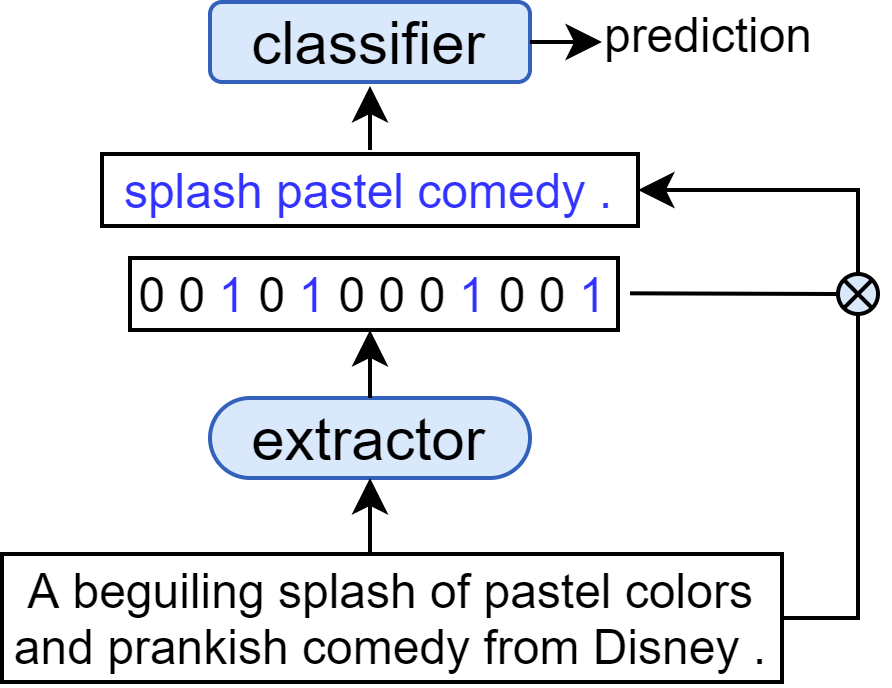}
	\caption{The overview of the model scheme, which consists of two components: extractor and classifier. Only the information of the selected subset can pass to the classifier.}
	\label{Ranker-Predictor}	
\end{figure}

The overview of our model is shown in Figure \ref{Ranker-Predictor}.
To guarantee that only the subset information is included in the classifier, we discretely select the words and pass words instead of the hidden states of the extractor to the classifier. 
Since gradients do not flow through discrete samples, we resort to HardKuma \citep{bastings-etal-2019-interpretable} to jointly train the model,
which gives support to binary outcomes.
HardKuma allows for setting the percentage of selected words and is proved more effective and stable than REINFORCE \citep{williams1992simple} in such scenarios. We set the selection ratio as 20\% for SST-2 and 40\% for MNLI because larger ratios will not cause further performance improvement. Finally, We get 85.6\% accuracy on SST-2 and 66.2/65.5  \% accuracy on MNLI-m/mm. 

~\\
\noindent
\textbf{Adversarial Attack Method \ \ }
We use TextFooler \citep{jin2020bert} to generate adversarial samples. 
We use the same settings to \citet{jin2020bert} to guarantee the semantical similarity of adversarial samples.
The only difference is that we search for samples with minimal similarity in the selected subset instead of the model prediction.
We guarantee that the model makes the same predictions, which is often used as the constraint for model reasoning in evaluation 3.
We generate adversarial samples with  10\% and 20\%  perturbation ratios. 

~\\
\noindent
\textbf{Results \ \ }
We use F1-score to compute the similarity score between subsets and report the Macro-averaging F1-score of the whole development set. 
A lower score is better, reflecting a larger difference in selected subsets.
Note that since some words in original samples are replaced with their synonyms in adversarial samples, synonyms are seen as identical to their original words when evaluating.
We evaluate all samples in the SST-2 development set and the first 1000 samples in MNLI-m/mm development sets.
The results are shown in Table \ref{result3}

\begin{table}[t!]
\centering
\begin{tabular}{|l|c|c|}
\hline
\textbf{Dataset/Ratio} & \ 10\%  \ &  \ 20\%  \  \\
\hline
SST-2 & 0.32 & 0.18 \\

MNLI-m/mm  & 0.43 / 0.52   &  0.37 / 0.43   \\
\hline
\end{tabular}
\caption{The similarity scores between selected subsets. \emph{Rato} refers to the perturbation ratio used to generate adversarial samples. }
\label{result3}
\end{table}

\begin{table}[t!]
\centering
\begin{tabular}{|l|c|c|}
\hline
\textbf{Dataset/Ratio} & \ 10\%  \ &  \ 20\%  \  \\
\hline
SST-2 & 0.32 & 0.51 \\

\hline
\end{tabular}
\caption{The similarity scores between selected subsets. \emph{Rato} refers to the perturbation ratio used to generate adversarial samples. }
\label{result3}
\end{table}

As shown in Table \ref{result3}, though semantically similar to the original samples and share the same model predictions, the adversarial samples can have subsets with low similarity to the original subset. Moreover, with a 10\% perturbation ratio, \textbf{31.8\%} of samples in SST-2 have an adversarial subset with none word overlap with the original subset. This result increases to \textbf{50.5\%} with a 20\% perturbation ratio.
With no overlap between the two subsets, there is no way we can hypothesis the adversarial samples share similar model reasoning to the original samples.

~\\
\noindent
\textbf{Summarization \ \ }
Though evaluation 3 seems reasonable, sharing similar semantics and the same model predictions is a weak constraint for similar model reasoning. Thus the change in attribution scores may come from different model reasoning instead of the instability of attribution methods.
Because of deep models' high sensitivity to adversarial samples, works resorting to evaluation 3 all get the same conclusion that existing attribution methods are fragile or unreliable.
We argue we should find a more strict constraint for model reasoning first instead of ignoring logic trap 3 and disproving attribution methods recklessly.

\section{Discussion}

\subsection{Attacking attribution methods by replacing the target model.}

Besides resorting to methods in evaluation 3, there are works \citep{jain2019attention, wang2020gradient, slack2020fooling} disprove the reliability of attribution methods by replacing the target model which attribution methods should work on.

For example, \citet{slack2020fooling} trains an adversarial classifier $e(x)$ to distinguish whether the inputs have been perturbed or not and then uses a different sub-model to process perturbed instances.
Specifically, if we want to attack the LOO method, we can build a loo set from the original dataset and train $e(x)$ in the following form:
\[
e(x) = \begin{cases}
f(x), & if \ x \in original \ set \\ 
\psi(x), & if \  x \in loo \  set
\end{cases}
\]
This way, $\psi(x)$, a model irrelevant to model predictions, is used when using LOO to calculate attribution scores, making generated attribution scores meaningless. \citet{slack2020fooling} assert that results of perturbation-based attribution methods such as LIME and SHAP \citep{lundberg2017unified} are easily attacked by their method.
Similarly, \citet{wang2020gradient} add an extra model to the original model, which has uniform outputs but large gradients for some particular tokens such as `CLS' in BERT. Since the extra model generates uniform outputs, it will not affect predictions of the original model.
However, the extra model's gradients will add to the original model and thus can confuse gradient-based attribution methods.

\subsection{Should We Use Attribution Methods in a Black-Box Way?}

The attack methods in Section 3.1 fool the attribution methods through designing a special structure and require 
attribution methods to be used in a black-box way.
In this setting, the attribution methods are easily attacked and generate meaningless results.
However, the question is: as a tool to help humans understand how deep models work, is it necessary to use attribution methods in a black-box way?
Take the linear model as an example. 
The linear model is regarded as a white-box model, and humans don't need attribution methods to understand how it works.
However, the understanding of a linear model is based on the analysis of its calculation process. Meanwhile, the deep model is regarded as a black-box model because its calculation process is too complicated to understand for humans, not because its calculation process is inaccessible.
Thus, we believe there are no compelling reasons to require attribution methods to be used in a black-box way.
The attacks in \citet{wang2020gradient, slack2020fooling} will fail when humans use attribution methods with knowing the model structures.

\subsection{Reducing the impact of proposed logic traps.}
Since logic traps in existing evaluation methods can cause an inaccurate evaluation,
we believe reducing the impact of these traps is the next question in the research field of post-hoc interpretations.
In this section, we provide some thoughts for reducing the impact of logic trap 3: 

\emph{The change in attribution scores may be because the model reasoning process is changed rather than the attribution method is unreliable.}

To reduce the impact of this logic trap, we should try to guarantee the similarity in model reasoning when processing semantically similar inputs.
In other words, we hope the target model used to test attribution methods more robustness to adversarial samples, which can be conducted through the following ways:

\begin{itemize}
\item[1] \textbf{Enhancing the target model.}
The success of adversarial attacks on deep models motivates efforts to defend against such attacks.
Thus, we can use these defense techniques, such as adversarial training \citep{tramer2017ensemble} and randomization \citep{xie2017mitigating}, to enhance the target model and make it more robustness to adversarial samples.

\item[2] \textbf{Excluding predictions with low confidence.}
The deep model will give a prediction for a sample regardless of whether knowing how to deal with it.
The randomness of reasoning increases with the uncertainty in model decisions \citep{bella2010calibration}.
Thus, we can guarantee the stability of model reasoning by excluding low-confident predictions.
For example, we can resorting to \textbf{Confidence Calibration techniques} \citep{guo2017calibration, seo2019learning}, which calculate confidence interval for a predicted response.

\end{itemize}

\subsection{Conclusions}

The proposed logic traps in existing evaluation methods have been ignored for a long time and negatively affected the research field. 
Though strictly accurate evaluation metrics for evaluating attribution methods might be a “unicorn” which will likely never be found, we should not just ignore these logic traps and draw conclusions recklessly.
With a clear statement and awareness of these logic traps, we should reduce the focus on improving performance under such unreliable evaluation systems and shift it to reducing the impact of proposed logic traps.
Moreover, other aspects of the research field should give rise to more attention, such as the applications of attribution scores (denoising data, improving the model performance, etc.) and proposing new explanation forms.

\section*{Acknowledgements}
This work is supported by the National Key Research and Development Program of China (No.2020AAA0106400), the National Natural Science Foundation of China (No.61922085, No.61976211, No.61906196). This work is also supported by the Key Research Program of the Chinese Academy of Sciences (Grant NO. ZDBS-SSW-JSC006), the  independent research project of National Laboratory of Pattern Recognition and in part by the Youth Innovation Promotion Association CAS.


\bibliography{anthology,custom}
\bibliographystyle{acl_natbib}

\appendix

\section{Experimental Details}
In this section, we provide the experimental details of our experiments.
Moreover, we will release our code and model within two months.

\subsection{Experiment 1}
\begin{figure}[htb]
	\centering
	\includegraphics[width=0.6\linewidth]{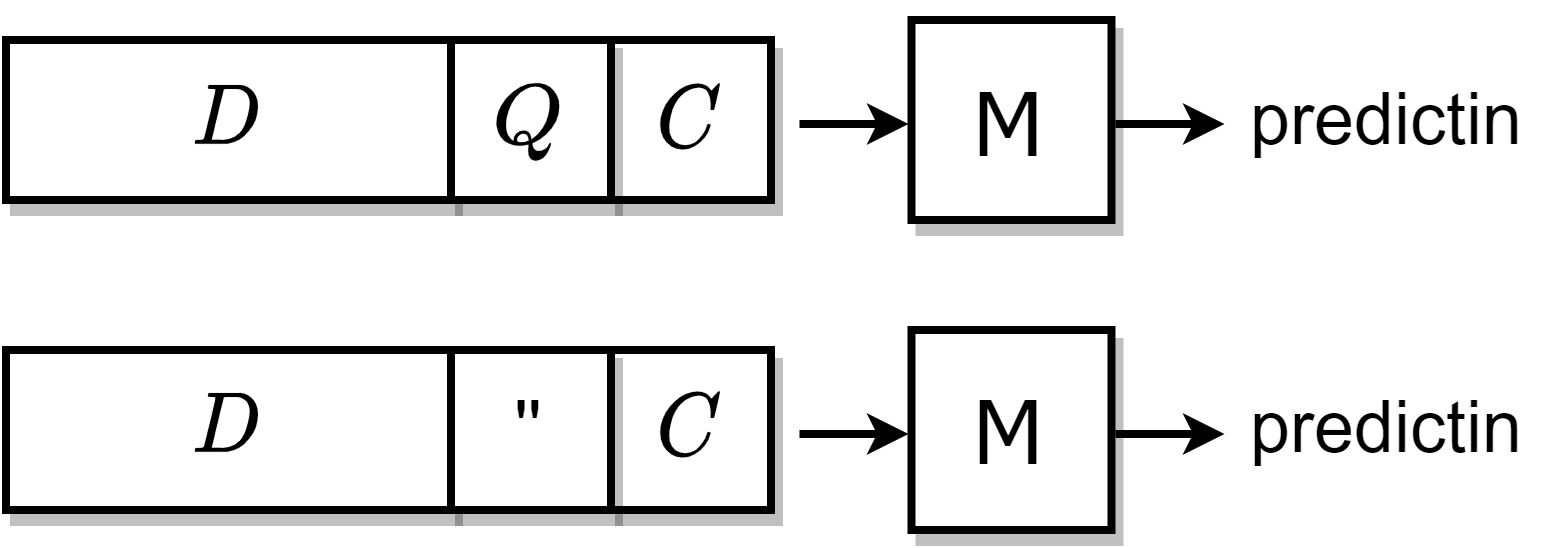}
	\caption{The overview of experiment 1.}
	\label{E1}	
\end{figure}

We merged dev-high and dev-middle sets as the development set.
As shown in Figure \ref{E1}, the document $D$, question $Q$, and one of the choices $C$ are concatenated together as the input of model, and we replace the development set questions with empty strings in our experiment.

\subsection{Experiment 2}

We use the tokenizer of BERT to split the sentence into words in experiment 2.
We modify 20\% of words in the sentences in experiment 2. Since the word number in a sentence is not necessarily a multiple of five, we need to choose between rounding up or down.
We use the same setting in code of HEDGE, i.e., rounding down.
Specifically, we modify one word when word number is smaller than five.

\subsection{Experiment 3}
\begin{figure}[htb]
	\centering
	\includegraphics[width=0.99\linewidth]{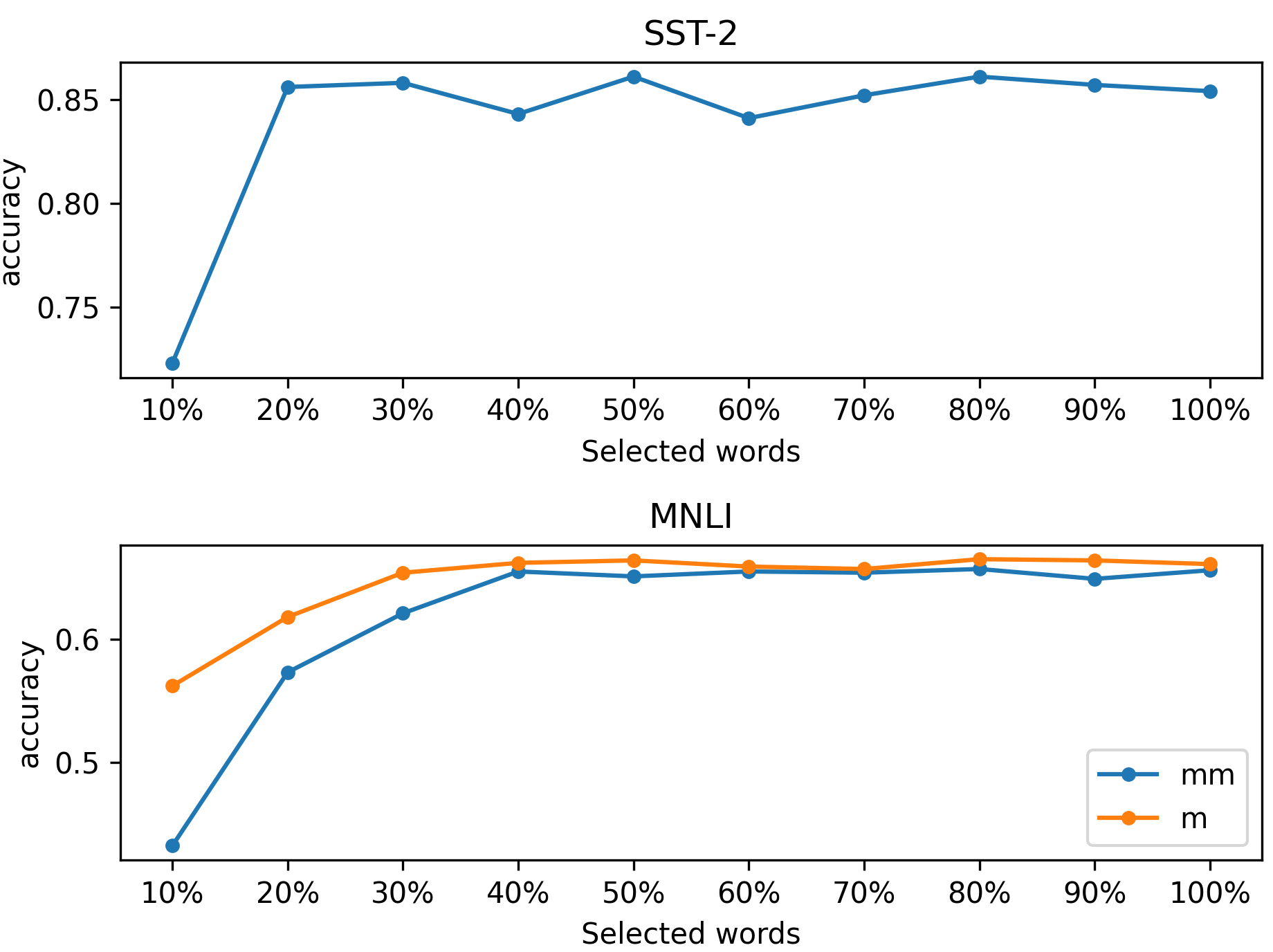}
	\caption{Model performance trained in different settings.}
	\label{E3}	
\end{figure}

Since HardKuma allows for setting the percentage of selected words, we first experiment with settings ranging from 10\% to 100\%.
The results are shown in Figure \ref{E3}. 
Under the premise of maintaining model performance, we choose the smallest setting (20\% setting for SST-2 and 40\% setting for MNLI).
We use beam search to find adversarial samples and set the maximum reserved sample number to 100.

\end{document}